\documentclass[10pt,twocolumn,letterpaper]{article}
\pdfoutput=1
\usepackage{cvpr}
\usepackage{times}
\usepackage{epsfig}
\usepackage{graphicx}
\usepackage{subfig}
\usepackage{float}
\usepackage{amsmath}
\usepackage{amssymb}
\usepackage{array}
\usepackage{enumitem}
\usepackage{multirow}
\def\signed #1{{\leavevmode\unskip\nobreak\hfil\penalty50\hskip2em
  \hbox{}\nobreak\hfil(#1)%
  \parfillskip=0pt \finalhyphendemerits=0 \endgraf}}
\newsavebox\mybox

\usepackage[lined,linesnumbered,ruled]{algorithm2e}

\SetCommentSty{mycommfont}

\usepackage{pifont}
\newcommand{\specialcell}[2][c]{\begin{tabular}[#1]{@{}c@{}}#2\end{tabular}}

\usepackage[pagebackref=true,breaklinks=true,letterpaper=true,colorlinks,bookmarks=false]{hyperref}

\cvprfinalcopy 

\def\cvprPaperID{3908} 

\begin{document}

\title{Explainable and Explicit Visual Reasoning over Scene Graphs}

\author{Jiaxin Shi$^{1}$\protect \thanks{The work was done when Jiaxin Shi was an intern at Nanyang Technological University.} ~~~~~~~~Hanwang Zhang$^{2}$   ~~~~~~~~Juanzi Li$^1$\\
$^1$Tsinghua University ~~~~
$^2$Nanyang Technological University \\
shijx12@gmail.com; hanwangzhang@ntu.edu.sg; lijuanzi@tsinghua.edu.cn
}

\maketitle

\begin{abstract}
We aim to dismantle the prevalent black-box neural architectures used in complex visual reasoning tasks, into the proposed eXplainable and eXplicit Neural Modules (XNMs), which advance beyond existing neural module networks towards using scene graphs --- objects as nodes and the pairwise relationships as edges --- for explainable and explicit reasoning with structured knowledge. XNMs allow us to pay more attention to teach machines how to ``think'', regardless of what they ``look''. As we will show in the paper, by using scene graphs as an inductive bias, 1) we can design XNMs in a concise and flexible fashion, i.e., XNMs merely consist of 4 meta-types, which significantly reduce the number of parameters by 10 to 100 times, and 2) we can explicitly trace the reasoning-flow in terms of graph attentions. XNMs are so generic that they support a wide range of scene graph implementations with various qualities. For example, when the graphs are detected perfectly, XNMs achieve 100\% accuracy on both CLEVR and CLEVR CoGenT, establishing an empirical performance upper-bound for visual reasoning; when the graphs are noisily detected from real-world images, XNMs are still robust to achieve a competitive 67.5\% accuracy on VQAv2.0, surpassing the popular bag-of-objects attention models without graph structures. 
\end{abstract}

\vspace{-0.4cm}
\section{Introduction}
\begin{figure}[t]
\includegraphics[width=\linewidth]{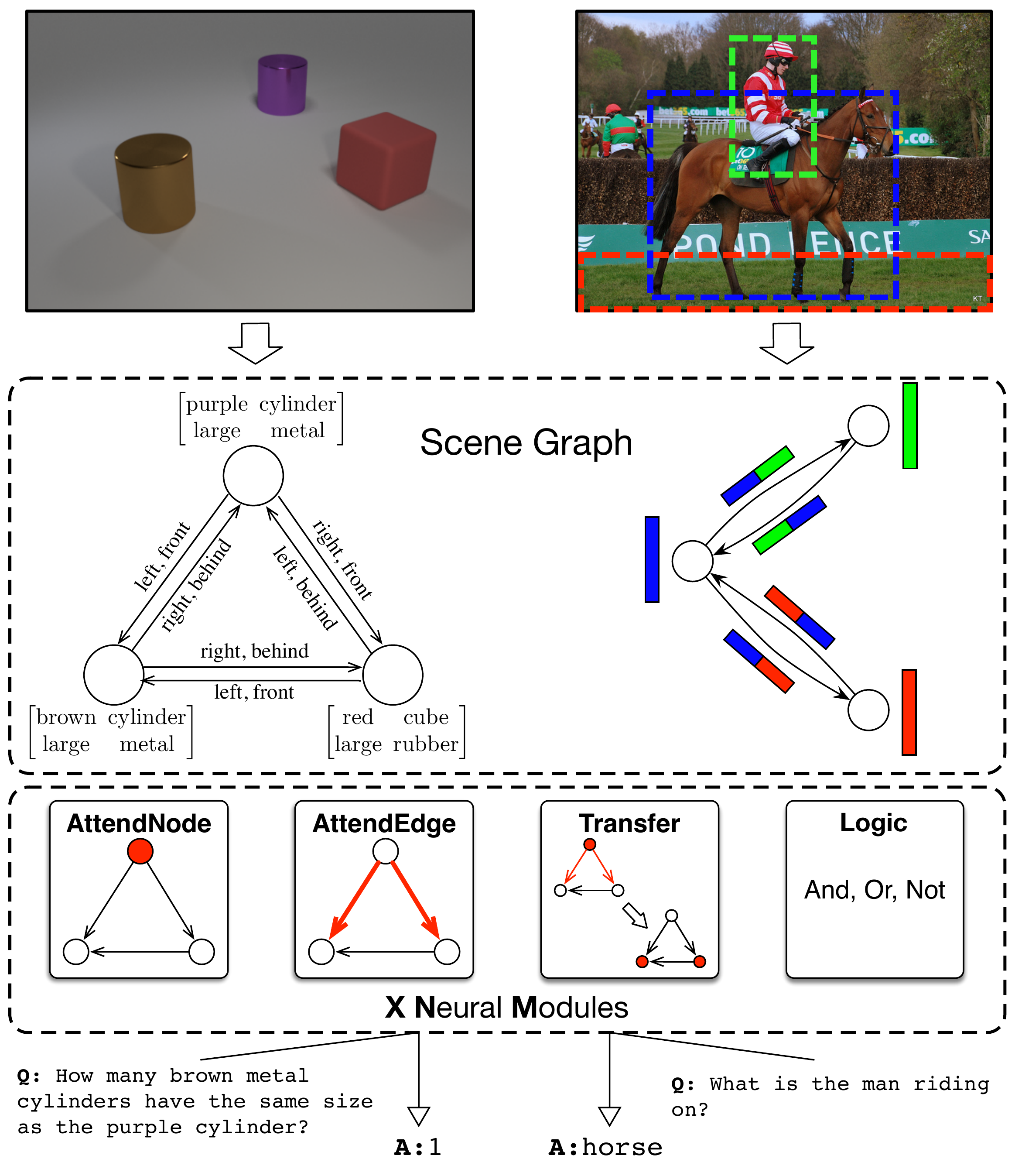}
\caption{The flowchart of using the proposed XNMs reasoning over scene graphs, which can be represented by detected one-hot class labels (left) or RoI feature vectors (colored bars on the right). Feature colors are consistent with the bounding box colors. XNMs have 4 meta-types. Red nodes or edges indicate attentive results. The final module assembly can be obtained by training an off-the-shelf sequence-to-sequence program generator~\cite{johnson2017inferring}.}
\label{fig:1}
\vspace{-0.5cm}
\end{figure}

The prosperity of A.I. --- mastering super-human skills in game playing~\cite{silver2016alphago}, speech recognition~\cite{amodei2016speech}, and image recognition~\cite{he2016deep,ren2015faster} --- is mainly attributed to the ``winning streak'' of \emph{connectionism}, more specifically, the deep neural networks~\cite{lecun2015deep}, over the ``old-school'' \emph{symbolism}, where their controversy can be dated back to the birth of A.I. in 1950s~\cite{minsky1991versus}. With massive training data and powerful computing resources, the key advantage of deep neural networks is the end-to-end design that generalizes to a large spectrum of domains, minimizing the human efforts in  domain-specific knowledge engineering. However, large gaps between human and machines can be still observed in ``high-level'' vision-language tasks such as visual Q\&A~\cite{antol2015vqa,goyal2017vqa2,johnson2017clevr}, which inherently requires composite reasoning (cf. Figure~\ref{fig:1}). In particular, recent studies show that the end-to-end models are easily optimized to learn the dataset ``shortcut bias'' but not reasoning~\cite{johnson2017clevr}.

Neural module networks (NMNs)~\cite{andreas2016neural,johnson2017clevr,hu2017learning,mascharka2018transparency,hu2018explainable,yi2018nsvqa} show a promising direction in conferring reasoning ability for the end-to-end design by learning to compose the networks on-demand from the language counterpart, which implies the logical compositions. Take the question \emph{``How many objects are left of the red cube?''} as an example, we can program the reasoning path into a composition of functional modules~\cite{mascharka2018transparency}: \texttt{Attend[cube]}, \texttt{Attend[red]}, \texttt{Relate[left]}, and \texttt{Count}, and then execute them with the input image.  
We attribute the success of NMNs to the eXplainable and eXplicit (dubbed \textbf{X}) language understanding.
By explicitly parsing the question into an explainable module assembly, NMNs effectively prevent the language-to-reasoning shortcut, which are frequent when using the implicit fused question representations~\cite{antol2015vqa,goyal2017vqa2} (\eg, the answer can be directly inferred according to certain language patterns).

However, the vision-to-reasoning shortcut still exists as an obstacle on the way of NMNs towards the real X visual reasoning. This is mainly because that the visual perception counterpart is still attached to reasoning~\cite{mascharka2018transparency}, which is inevitably biased to certain vision patterns. For example, on the CLEVR CoGenT task, which provides novel object attributes to test the model's generalization ability (\eg, cubes are blue in the training set but red in the test set), we observe significant performance drop of existing NMNs~\cite{johnson2017inferring,mascharka2018transparency} (\eg, red cubes in the test set cannot be recognized as ``cube''). 
Besides, the reusability of the current module design is limited. For example, the network structure of the \texttt{Relate} module in~\cite{mascharka2018transparency} must be carefully designed using a series of dilated convolutions to achieve good performance. Therefore, how to design a complete inventory of X modules is still an tricky engineering. 

In this paper, we advance NMN towards X visual reasoning by using the proposed e\textbf{X}plainable and e\textbf{X}plicit \textbf{N}eural \textbf{M}odules (XNMs) reasoning over \emph{scene graphs}. By doing this, we can insulate the ``low-level'' visual perception from the modules, and thus can prevent reasoning shortcut of both language and vision counterpart. As illustrated in Figure~\ref{fig:1}, a scene graph is the knowledge representation of a visual input, where the nodes are the entities (\eg, cylinder, horse) and the edges are the relationships between entities (\eg, left, ride). In particular, we note that scene graph detection \textit{per se} is still a challenging task in computer vision~\cite{zellers2018motifs}, therefore, we allow XNMs to accept scene graphs with different detection qualities. For example, the left-hand side of Figure~\ref{fig:1} is one extreme when the visual scene is clean and closed-vocabulary, \eg, in CLEVR~\cite{johnson2017clevr}, we can have almost perfect scene graphs where the nodes and edges can be represented by one-hot class labels; the right-hand side shows another extreme when the scene is cluttered and open-vocabulary in practice, the best we have might be merely a set of object proposals. Then, the nodes are RoI features and the edges are their concatenations. 

Thanks to scene graphs, our XNMs only have 4 meta-types: 1) \texttt{AttendNode}, finding the queried entities, 2) \texttt{AttendEdge}, finding the queried relationships, 3) \texttt{Transfer}, transforming the node attentions along the attentive edges, and 4) \texttt{Logic}, performing basic logical operations on attention maps. All types are fully X as their outputs are pure graph attentions that are  easily traceable and visible. 
Moreover, these meta modules are only specific to the generic graph structures, and are highly reusable to constitute different composite modules for more complex functions.
For example, we do not need to carefully design the internal implementation details for the module \texttt{Relate} as in~\cite{mascharka2018transparency}; instead, we only need to combine \texttt{AttendEdge} and \texttt{Transfer} in XNMs.

We conduct extensive experiments \footnote{Our codes are public at \url{https://github.com/shijx12/XNM-Net}}
on two visual Q\&A benchmarks and demonstrate the following advantages of using XNMs reasoning over scene graphs:
\begin{enumerate}[leftmargin=.2in,itemsep=0pt,parsep=0pt,topsep=1pt, partopsep=0pt]
    \item We achieve 100\% accuracy by using the ground-truth scene graphs and programs on both CLEVR~\cite{johnson2017clevr} and CLEVR-CoGent, revealing the performance upper-bound of XNMs, and the benefits of disentangling ``high-level'' reasoning from ``low-level'' perception.
    \item Our network requires significantly less parameters while achieves better performance than previous state-of-the-art neural module networks, due to the conciseness and high-reusability of XNMs.
    \item XNMs are flexible to different graph qualities, \eg,  it achieves competitive accuracy on VQAv2.0~\cite{goyal2017vqa2} when scene graphs are noisily detected.
    \item We show qualitative results to demonstrate that our XNMs reasoning is highly explainable and explicit.
\end{enumerate}

\vspace{-0.2cm}
\section{Related Work}
\vspace{-0.1cm}
\textbf{Visual Reasoning.}
It is the process of analyzing visual information and solving problems based on it. The most representative benchmark of visual reasoning is CLEVR~\cite{johnson2017clevr}, a diagnostic visual Q\&A dataset for compositional language and elementary visual reasoning. The majority of existing methods on CLEVR can be categorized into two families: 
1) holistic approaches~\cite{johnson2017clevr,santoro2017simple,perez2017film,hudson2018compositional}, which embed both the image and question into a feature space and infer the answer by feature fusion;
2) neural module approaches~\cite{andreas2016neural,hu2017learning,johnson2017inferring,mascharka2018transparency,hu2018explainable,yi2018nsvqa}, which first parse the question into a program assembly of neural modules, and then execute the modules over the image features for visual reasoning. Our XNM belongs to the second one but replaces the visual feature input with scene graphs.

\textbf{Neural Module Networks.} They dismantle a complex question into several sub-tasks, which are easier to answer and more transparent to follow the intermediate outputs. Modules are pre-defined neural networks that implement the corresponding functions of sub-tasks, and then are assembled into a layout dynamically, usually by a sequence-to-sequence program generator given the input question. The assembled program is finally executed for answer prediction~\cite{hu2017learning,johnson2017inferring,mascharka2018transparency}. In particular, the program generator is trained based on the human annotations of desired layout or with the help of reinforcement learning due to the non-differentiability of layout selection. Recently, Hu \etal~\cite{hu2018explainable} proposed StackNMN, which replaces the hard-layout with soft and continuous module layout and performs well even without layout annotations at all. Our XNM experiments on VQAv2.0 follows their soft-program generator. 

Recently, NS-VQA~\cite{yi2018nsvqa} firstly built the reasoning over the object-level structural scene representation, improving the accuracy on CLEVR from the previous state-of-the-art 99.1\%~\cite{mascharka2018transparency} to an almost perfect 99.8\%. Their scene structure consists of objects with detected labels, but lacked the relationships between objects, which limited its application on real-world datasets such as VQAv2.0~\cite{goyal2017vqa2}. In this paper, we propose a much more generic framework for visual reasoning over scene graphs, including object nodes and relationship edges represented by either labels or visual features. Our scene graph is more flexible and more powerful than the table structure of NS-VQA.

\textbf{Scene Graphs.} This task is to produce graph representations of images in terms of objects and their relationships. Scene graphs have been shown effective in boosting several vision-language tasks~\cite{johnson2015image,teney2017graph,yin2017obj2text,chen2018scene}. To the best of our knowledge, we are the first to design neural module networks that can reason over scene graphs. However, scene graph detection is far from satisfactory compared to object detection~\cite{xu2017scene,zellers2018motifs,li2018factorizable}. To this end, our scene graph implementation also supports cluttered and open-vocabulary in real-world scene graph detection, where the nodes are merely RoI features and the edges are their concatenations.

\begin{figure*}[ht]
\includegraphics[width=\linewidth]{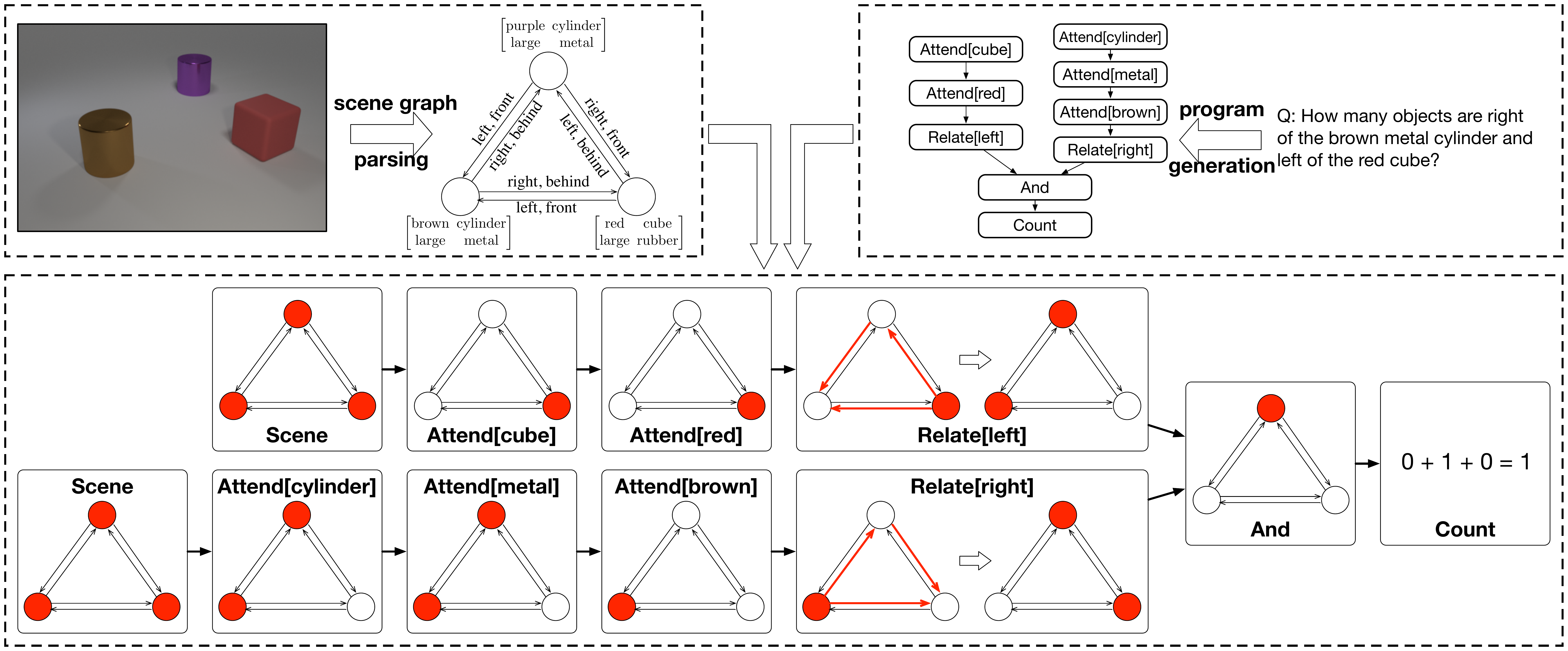}
\caption{To answer a question about an image, we need to 1) parse the image into a scene graph, 2) parse the question into a module program, and 3) reasoning over the scene graph. Here, we show the reasoning details of an example from CLEVR. The nodes and edges in red are attended. \texttt{Scene} is a dummy placeholder module that attends all nodes. All intermediate steps of our XNMs are explainable and explicit.}
\label{fig:2}
\vspace{-0.4cm}
\end{figure*}

\vspace{-0.2cm}
\section{Approach}
We build our neural module network over scene graphs to tackle the visual reasoning challenge.
As shown in Figure~\ref{fig:2}, given an input image and a question, we first parse the image into a scene graph and parse the question into a module program, and then execute the program over the scene graph. In this paper, we propose a set of generic base modules that can conduct reasoning over scene graphs --- e\textbf{X}plainable and e\textbf{X}plicit \textbf{N}eural \textbf{M}odules (XNMs) --- as the reasoning building blocks. We can easily assemble these XNMs to form more complex modules under specific scenarios. Besides, our XNMs are totally attention-based, making all the intermediate reasoning steps transparent.

\vspace{-0.1cm}
\subsection{Scene Graph Representations}
\vspace{-0.1cm}
We formulate the scene graph of an image as $(\mathcal{V}, \mathcal{E})$, where $\mathcal{V}=\{\mathbf{v}_1, \cdots, \mathbf{v}_N\}$ are graph nodes corresponding to $N$ detected objects, and $\mathbf{v}_i$ denotes the feature representation of the $i$-th object.
$\mathcal{E}=\{\mathbf{e}_{ij} | i,j=1,\cdots,N\}$ are graph edges corresponding to relations between each object pairs, and $\mathbf{e}_{ij}$ denotes the feature representation of the relation from object $i$ to object $j$ (Note that edges are directed).

Our XNMs are generic for scene graphs of different quality levels of detection.
We consider two extreme settings in this paper.
The first is the ground-truth scene graph with labels, denoted by \textbf{GT}, that is, using ground-truth objects as nodes, ground-truth object label embeddings as node features, and ground-truth relation label embeddings as edge features.
In this setting, scene graphs are annotated with fixed-vocabulary object labels and relationship labels, \eg, defined in CLEVR dataset~\cite{johnson2017clevr}.
We collect all the $C$ labels into a dictionary, and use an embedding matrix $\mathbf{D} \in \mathbb{R}^{C \times d}$ to map a label into a $d$-dimensional vector.
We represent the nodes and edges using the concatenation of their corresponding label embeddings.

The second setting is totally detected and label-agnostic, denoted by \textbf{Det}, that is, using detected objects as nodes, RoI visual features as node features, and the fusion of two node features as the edge features.
For example, the edge features can be represented by concatenating the two related node features, \ie, $\mathbf{e}_{ij} = \big[ \mathbf{v}_i; \mathbf{v}_j \big]$. As an another example, in CLEVR where the edges are only about spatial relationships, we use the difference between detected coordinates of object pairs as the edge embedding. More details are in Section~\ref{sec:exp}.

We use the GT setting to demonstrate the performance upper-bound of our approach when a perfect scene graph detector is available along with the rapid development of visual recognition, and use the Det setting to demonstrate the practicality in open domains.

\subsection{X Neural Modules}
As shown in Figure~\ref{fig:1}, our XNMs have four meta-types and are totally attention-based. 
We denote the node attention weight vector by $\mathbf{a} \in [0,1]^{N}$ and the weight of the $i$-th node by $a_i$.
The edge attention weight matrix is denoted by $\mathbf{W} \in [0,1]^{N \times N}$, where $W_{ij}$ represents the weight of edge from node $i$ to node $j$.

\textbf{\texttt{AttendNode[query]}.}
This most basic and intuitive operation is to find the relevant objects given an input query (\eg, find all [``cubes'']).
For the purpose of semantic computation, we first encode the query into a vector $\mathbf{q}$.
This X module takes the query vector as input, and produces the node attention vector by the following function:
\begin{equation}\label{eq:1}
    \mathbf{a} = f(\mathcal{V}, \mathbf{q}).
\end{equation}
The implementation of $f$ is designed according to a specific scene graph representation, as long as $f$ is differentiable and $\mathrm{range}(f)=[0,1]$.

\textbf{\texttt{AttendEdge[query]}.}
Though object attention is a widely-used mechanism for better visual understanding, it is unable to capture the interaction between objects and thus is weak in the complex visual reasoning~\cite{zhang2018count}.
This X module aims to find the relevant edges given an input query (\eg, find all edges that are [``left'']).
After encoding the query into $\mathbf{q}$, we compute the edge attention matrix by the following function:
\begin{equation}\label{eq:2}
    \mathbf{W} = g(\mathcal{E}, \mathbf{q}),
\end{equation}
where $g$ is defined according to a specific scene graph representation, as long as $g$ is differentiable and $\mathrm{range}(g)=[0,1]$.

\textbf{\texttt{Transfer}.}
With the node attention vector $\mathbf{a}$ and the edge attention matrix $\mathbf{W}$, we can transfer the node weights along the attentive relations to find new objects (\eg, find objects that are [``left''] to the [``cube'']).
Thanks to the graph structure, to obtain the updated node attention $\mathbf{a}'$, we merely need to perform a simple matrix multiplication:
\begin{equation}\label{eq:transfer}
    \mathbf{a}' = \mathrm{norm} (\mathbf{W}^\top \mathbf{a}),
\end{equation}
where $\mathrm{norm}$ assert the values in $[0,1]$ by dividing the maximum value if any entry exceeds $1$.
Here, $W_{ij}$ indicates how many weights will flow from object $i$ to object $j$, and $a'_i = \sum_{j=1}^{N} W_{ji} a_j$ is the total received weights of object $i$.
This module reallocates node attention in an efficient and fully-differentiable manner.

\textbf{Logic.}
Logical operations are crucial in complex reasoning cases.
In XNM, logical operations are performed on one or more attention weights to produce a new attention.
We define three logical X modules: \textbf{\texttt{And}}, \textbf{\texttt{Or}}, and \textbf{\texttt{Not}}.
Without loss of generality, we discuss all these logical modules on node attention vectors, and the extension to edge attention is similar.
The \texttt{And} and \texttt{Or} modules are binary, that is, take two attentions as inputs, while the \texttt{Not} module is unary.
The implementation of these logical X modules are as follows:

\vspace{-0.4cm}
\begin{equation}
\begin{aligned}
    &\texttt{And}(\mathbf{a}^1, \mathbf{a}^2) = \text{min}(\mathbf{a}^1,\mathbf{a}^2), \texttt{Not}(\mathbf{a})  = 1 - \mathbf{a},\\ &\texttt{Or}(\mathbf{a}^1,  \mathbf{a}^2) = \text{max}(\mathbf{a}^1, \mathbf{a}^2).
\end{aligned}
\end{equation}

These four meta-types of XNMs constitute the base of our graph reasoning.
They are explicitly executed on attention maps, and all intermediate results are explainable.
Besides, these X modules are totally differentiable.
We can flexibly assemble them into composite modules for more complex functions, which can be still trained end-to-end.

\subsection{Implementations}
To apply XNMs in practice, we need to consider these questions:
\textbf{(1)} How to implement the attention functions $f,g$ in Eq.~\eqref{eq:1} and Eq.~\eqref{eq:2}?
\textbf{(2)} How to compose our X modules into composite reasoning modules? \textbf{(3)} How to predict the answer according to the attentive results?
\textbf{(4)} How to parse the input question to an executable module program?

\vspace{-0.2cm}
\subsubsection{Attention Functions}
We use different attention functions for different scene graph settings.
In the GT setting, as annotated labels are mostly mutually exclusive (\eg, ``red'' and ``green''), we compute the node attention using the softmax function over the label space.
Specifically, given a query vector $\mathbf{q} \in \mathbb{R}^d$, we first compute its attention distribution over all labels by $\mathbf{b} = \mathrm{softmax}(\mathbf{D} \cdot \mathbf{q})$, where $\mathrm{length}(\mathbf{b})=C$ and $b_c$ represents the weight of the $c$-th label.
Then we capture the node and edge attention by summing up corresponding label weights:
\begin{equation}\label{eq:softmax-att}
    a_i = f(\mathcal{V}, \mathbf{q})_i = \sum_{c \in \mathcal{C}_i} b_c,~~
    W_{ij} = g(\mathcal{E}, \mathbf{q})_{ij} = \sum_{c \in \mathcal{C}_{ij}} b_c,
\end{equation}
where $\mathcal{C}_i$ and $\mathcal{C}_{ij}$ denote the (multi-) labels of node $i$ and edge $ij$ respectively.

In the Det setting, we use the sigmoid function to compute the attention weights.
Given the query $\mathbf{q} \in \mathbb{R}^d$, the node and edge attentions are:
\begin{equation}\label{eq:sigmoid-att}
\begin{aligned}
    a_i &= f(\mathcal{V}, \mathbf{q})_i = \mathrm{sigmoid}\left(\textrm{MLP}(\mathbf{v}_i)^\top \mathbf{q}\right),\\
    W_{ij} &= g(\mathcal{E}, \mathbf{q})_{ij} = \mathrm{sigmoid}\left(\textrm{MLP}(\mathbf{e}_{ij})^\top \mathbf{q}\right),
\end{aligned}
\end{equation}
where the MLP maps $\mathbf{v}_i$ and $\mathbf{e}_{ij}$ to the dimension $d$.

\vspace{-0.3cm}
\subsubsection{Composite Reasoning Modules}
We list our composite reasoning modules and their implementations (\ie, how they are composed by basic X modules) in the top section of Table~\ref{tab:modules}. For example, \texttt{Same} module is to find other objects that have the same attribute value as the input objects (\eg, find other objects with the same [``color'']).  In particular, \texttt{Describe} used in \texttt{Same} is to obtain the corresponding attribute value (\eg, describe one object's [``color'']), and will be introduced in the following section.

\vspace{-0.3cm}
\subsubsection{Feature Output Modules}
Besides the above reasoning modules, we also need another kind of modules to map the intermediate attention to a hidden embedding $\mathbf{h}$ for feature representation, which is fed into a softmax layer to predict the final answer, or into some modules for further reasoning.
We list our output modules in the bottom section of Table~\ref{tab:modules}.
\texttt{Exist} and \texttt{Count} sum up the node attention weights to answer yes/no and counting questions.
\texttt{Compare} is for attribute or number comparisons, which takes two hidden features as inputs.
\textbf{\texttt{Describe[query]}} is to transform the attentive node features to an embedding that describes the specified attribute value (\eg, what is the [``color''] of attended objects).

To implement the \texttt{Describe} module, we first obtain the ``raw'' attentive node feature by
\vspace{-0.2cm}
\begin{equation}\label{for:barv}
    \bar{\mathbf{v}} = \sum_{i=1}^N a_i \mathbf{v}_i \Big/ \sum_{i=1}^N a_i,
\end{equation}
and then project it into several ``fine-grained'' sub-spaces --- describing different attribute aspects such as color and shape --- using different transformation matrices. 
Specifically, we define $K$ projection matrices $\mathbf{M}_1, \cdots, \mathbf{M}_K$ to map $\bar{\mathbf{v}}$ into different aspects (\eg, $\mathbf{M}_1 \bar{\mathbf{v}}$ represents the color, $\mathbf{M}_2  \bar{\mathbf{v}}$ represents the shape, \etc), where $K$ is a hyper-parameter related to the specific scene graph vocabulary.
The output feature is computed by
\vspace{-0.2cm}
\begin{equation}\label{eq:query}
    \texttt{Describe}(\mathbf{a}, \mathbf{q}) = \sum_{k=1}^{K} c_k (\mathbf{M}_k \bar{\mathbf{v}}),
\vspace{-0.1cm}
\end{equation}
where $\mathbf{c} = \textrm{Softmax}(\textrm{MLP}(\mathbf{q}))$ represents a probability distribution over these $K$ aspects, and $c_k$ denotes the $k$-th probability.
The mapping matrixes can be learned end-to-end automatically.

\begin{table}[h]
    \caption{Our composite modules (the top section) and output modules (the bottom section). $\mathrm{MLP}()$ consists of several linear and ReLU layers.}
    \centering
    \footnotesize
    \begin{tabular}{|c|c|c|}
    \hline
    Modules & In $\to$ Out & Implementation \\
    \hline
    \texttt{Intersect}    & $\mathbf{a}^1, \mathbf{a}^2 \to \mathbf{a}'$       & $\texttt{And}(\mathbf{a}^1, \mathbf{a}^2)$ \\
    \texttt{Union}     & $\mathbf{a}^1, \mathbf{a}^2 \to \mathbf{a}'$          & $\texttt{Or}(\mathbf{a}^1, \mathbf{a}^2)$ \\
    \texttt{Filter}     & $\mathbf{a}$, $\mathbf{q} \to \mathbf{a}'$     & $\texttt{And}(\mathbf{a},  \texttt{AttendNode}(\mathbf{q}))$ \\
    \texttt{Same}       & $\mathbf{a}$, $\mathbf{q} \to \mathbf{a}'$     & {\fontsize{7}{7}\selectfont $\texttt{Filter}(\texttt{Not}(\mathbf{a}), \texttt{Describe}(\mathbf{a}, \mathbf{q}))$} \\
    \texttt{Relate}     & $\mathbf{a}$, $\mathbf{q} \to \mathbf{a}'$     & {\fontsize{7.7}{7.7}\selectfont $\texttt{Transfer}(\mathbf{a}, \texttt{AttendEdge}(\mathbf{q}))$} \\
    \hline
    \hline
    \specialcell[c]{\texttt{Exist} \\ \texttt{Count}}     & $\mathbf{a} \to \mathbf{h}$	       & $\mathrm{MLP}(\sum_i a_i)$\\
    \texttt{Compare}     & $\mathbf{h}^1, \mathbf{h}^2 \to \mathbf{h}'$    & $\mathrm{MLP}(\mathbf{h}^1 - \mathbf{h}^2)$ \\
    \texttt{Describe}  & $\mathbf{a}$, $\mathbf{q} \to \mathbf{h}$    & Eq.~\eqref{eq:query}\\
    
    \hline
    \end{tabular}
    \label{tab:modules}
\vspace{-0.4cm}
\end{table}

\begin{table*}[ht]
    \caption{Comparisons between neural module networks on the CLEVR dataset. Top section: results of the official test set; Bottom section: results of the validation set (we can only evaluate our GT setting on the validation set since the annotations of the test set are not public~\cite{johnson2017clevr}). The program option ``scratch'' means totally without program annotations, ``supervised'' means using trained end-to-end parser, and ``GT'' means using ground-truth programs. Our reasoning modules are composed with highly-reusable X modules, leading to a very small number of parameters. Using the ground-truth scene graphs and programs, we can achieve a perfect reasoning on all kinds of questions.}
    \centering
    \small
    \begin{tabular}{|c|c|c|c||c|c|c|c|c|c|}
    \hline
    Method                          & Program & \#Modules & \#Param. & Overall  & Count & \specialcell[c]{Compare\\Numbers} & Exist & \specialcell[c]{Query\\Attribute} & \specialcell[c]{Compare\\Attribute} \\
    \hline
    Human~\cite{johnson2017clevr}   &   -          &   -   &   -       &   92.6  & 86.7  &     86.4    &   96.6   &   95.0   &   96.0 \\
    N2NMN~\cite{hu2017learning}     &   scratch    &   12  &   -       &   69.0  & -    & -    & -    & -    & -    \\
    N2NMN~\cite{hu2017learning}     &   supervised  &   12  &   -       &   83.7  & -    & -    & -    & -    & -    \\
    PG+EE~\cite{johnson2017inferring} &   supervised  &   39  &   40.4M   &   96.9  & 92.7 & 98.7 & 97.1 & 98.1 & 98.9 \\
    TbD-net~\cite{mascharka2018transparency} & supervised &   39  &   115M    &   \textbf{99.1}  & \textbf{97.6} & \textbf{99.4} & \textbf{99.2} & \textbf{99.5} & \textbf{99.6} \\
    StackNMN~\cite{hu2018explainable} &   scratch  &   9   &   7.32M   &      93.0  & -    & -    & -    & -    & -    \\
    StackNMN~\cite{hu2018explainable} &   supervised  &   9   &   7.32M   &   96.5  & -    & -    & -    & -    & -    \\
    XNM-Det  &   supervised  &   12  &   \textbf{0.55M}   &   97.7 & 96.0 & 98.0 & 98.7 & 98.4 & 97.6 \\
    \hline
    \hline
    NS-VQA~\cite{yi2018nsvqa}       &   supervised  &   12  &   -   &   99.8  & 99.7    & 99.9    & 99.9    & 99.8    & 99.8 \\
    XNM-Det  &   supervised  &   12  &   0.55M   &   97.8 & 96.0 & 98.1 & 98.6 & 98.7 & 97.8 \\
    XNM-Det  &   GT          &   12  &   0.55M   &   97.9 & 96.2 & 98.1 & 98.8 & 98.7 & 97.8 \\
    XNM-GT  &   supervised  &   12  &   0.22M   &   99.9 & 99.9 & 99.9 & 99.8 & 99.8 & 99.9 \\
    XNM-GT  &   GT          &   12  &   \textbf{0.22M}   &   \textbf{100}  & \textbf{100}  & \textbf{100}  & \textbf{100}  & \textbf{100}  & \textbf{100}  \\
    \hline
    \end{tabular}
    \label{tab:clevr-results}
\vspace{-0.3cm}
\end{table*}

\vspace{-0.2cm}
\subsubsection{Program Generation \& Training}
For datasets that have ground-truth program annotations (\eg, CLEVR), we directly learn an LSTM sequence-to-sequence model~\cite{sutskever2014sequence} to convert the word sequence into the module program.
However, there is no layout annotations in most real-world datasets (\eg, VQAv2.0). 
In this case, following StackNMN~\cite{hu2018explainable}, we make soft module selection with a differentiable stack structure.
Please refer to their papers for more details.

We feed our output features from modules (cf. Table~\ref{tab:modules}) into a softmax layer for the answer prediction. We use the cross entropy loss between our predicted answers and ground-truth answers to train our XNMs.

\vspace{-0.2cm}
\section{Experiments}\label{sec:exp}
\vspace{-0.1cm}
\subsection{CLEVR}
\vspace{-0.1cm}

\textbf{Settings.}
The CLEVR dataset~\cite{johnson2017clevr} is a synthetic diagnostic dataset that tests a range of visual reasoning abilities.
In CLEVR, images are annotated with ground-truth object positions and labels, and questions are represented as functional programs that consists of 13 kinds of modules.
Except the ``Unique'' module, which does not have actual operation, all the remaining 12 modules can correspond to our modules in Table~\ref{tab:modules}.
CLEVR modules ``Equal\_attribute'', ``Equal\_integer'', ``Greater\_than'' and ``Less\_than'' have the same implementation as our \texttt{Compare}, but with different parameters.
There are $4$ attribute categories in CLEVR, so we set the number of mapping matrixes $K=4$.

We reused the trained sequence-to-sequence program generator of~\cite{johnson2017inferring,mascharka2018transparency}, which uses prefix-order traversal to convert the program trees to sequences.
Note that their modules are bundled with input, \eg, they regard Filter[red] and Filter[green] as two different modules.
This will cause serious sparseness in the real-world case.
We used their program generator, but unpack the module and the input (\eg, Filter[red] and Filter[green] are the same module with different input query).

In the GT setting, we performed reasoning over the ground-truth scene graphs.
In the Det setting, we built the scene graphs by detecting objects and using RoI features as node embeddings and the differences between detected coordinates as edge embeddings.
Since CLEVR does not provide the bounding box or segmentation annotations of objects, it is hard to directly train an object detector. NS-VQA~\cite{yi2018nsvqa} trained a Mask R-CNN~\cite{he2017mask} for object segmentation by ``hacking'' the rendering process~\cite{johnson2017clevr}, which could perform very well due to the simplicity of visual scenes of CLEVR. However, as we expected to explore X modules in a noisier case, we chose the trained attention modules of TbD-net~\cite{mascharka2018transparency} as our object detector.
Specifically, we enumerated all possible combinations of object attributes (\eg, red, cube, metal, large), and tried to find corresponding objects using their attention modules (\eg, intersection of the output mask of Attend[red], Attend[cube], Attend[metal] and Attend[large], and then regarded each clique as a single object).
The detected results have some frequent mistakes, such as inaccurate position, wrongly merged nodes (two adjacent objects with the same attribute values are recognized as one). These detection noises allow us to test whether our XNMs are robust enough.

\textbf{Goals.}
We expect to answer the following questions according to the CLEVR experiments:
\textbf{Q1:} What is the upper bound of our X reasoning when both the vision and language perceptions are perfect?
\textbf{Q2:} Are our XNMs robust for noisy detected scene graphs and parsed programs?
\textbf{Q3:} What are the parameter and data efficiency, and the convergence speed of XNMs?
\textbf{Q4:} How is the explainability of XNMs?

\textbf{Results.}
Experimental results are listed in Table~\ref{tab:clevr-results}.
\textbf{A1:}
When using the ground-truth scene graphs and programs, we can achieve 100\% accuracy, indicating an inspiring upper-bound of visual reasoning.
By disentangling ``high-level'' reasoning from ``low-level'' perception and using XNMs, we may eventually conquer the visual reasoning challenge with the rapid development of visual recognition.

\textbf{A2:}
With noisy detected scene graphs, we can still achieve a competitive 97.9\% accuracy using the ground-truth programs, indicating that our X reasoning are robust to different quality levels of scene graphs.
When replacing the ground-truth programs with parsed programs, the accuracy drops by 0.1\% in both GT and Det settings, which is caused by minor errors of the program parser.

\textbf{A3:}
Due to the conciseness and high-reusability of X modules, our model requires significantly less parameters than existing models. 
Our GT setting only needs about 0.22M parameters, taking about 500MB memory with batch size of 128, while
PG+EE~\cite{johnson2017inferring} and TbD-net~\cite{mascharka2018transparency} bundle modules and inputs together, leading to a large number of modules and parameters.

\begin{figure}[h]
\includegraphics[width=\linewidth]{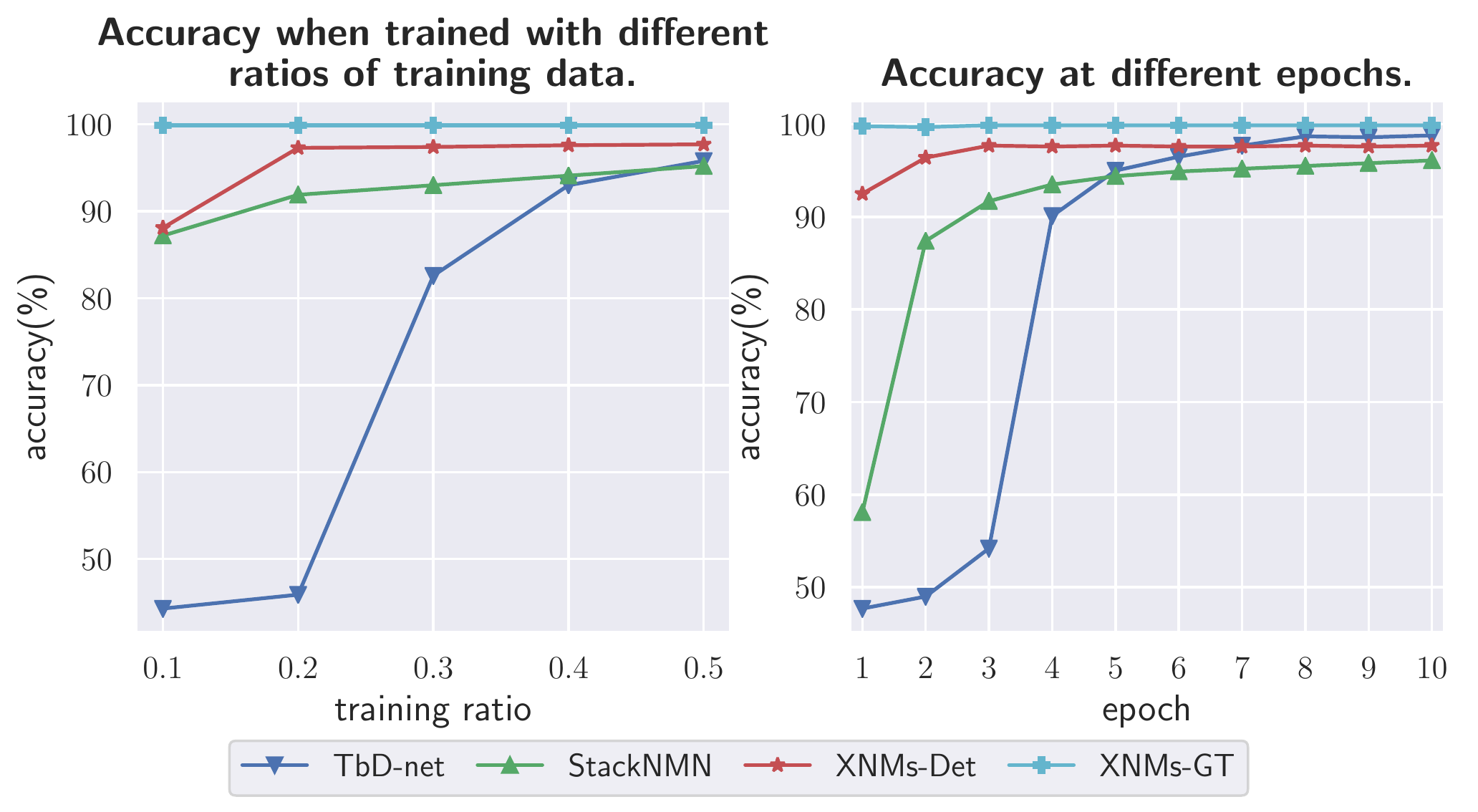}
\caption{Comparison of data efficiency and convergence speed.}
\label{fig:efficiency}
\vspace{-0.2cm}
\end{figure}

To explore the data efficiency, we trained our model with a partial training set and evaluated on the complete validation set.
Results are displayed in the left part of Figure~\ref{fig:efficiency}.
We can see that our model performs much better than other baselines when the training set is small.
Especially, our GT setting can still achieve a 100\% accuracy even with only 10\% training data.
The right part shows the accuracy at each training epoch.
We can see that our X reasoning converges very fast.

\begin{figure*}[t]
\centering
\includegraphics[width=\linewidth]{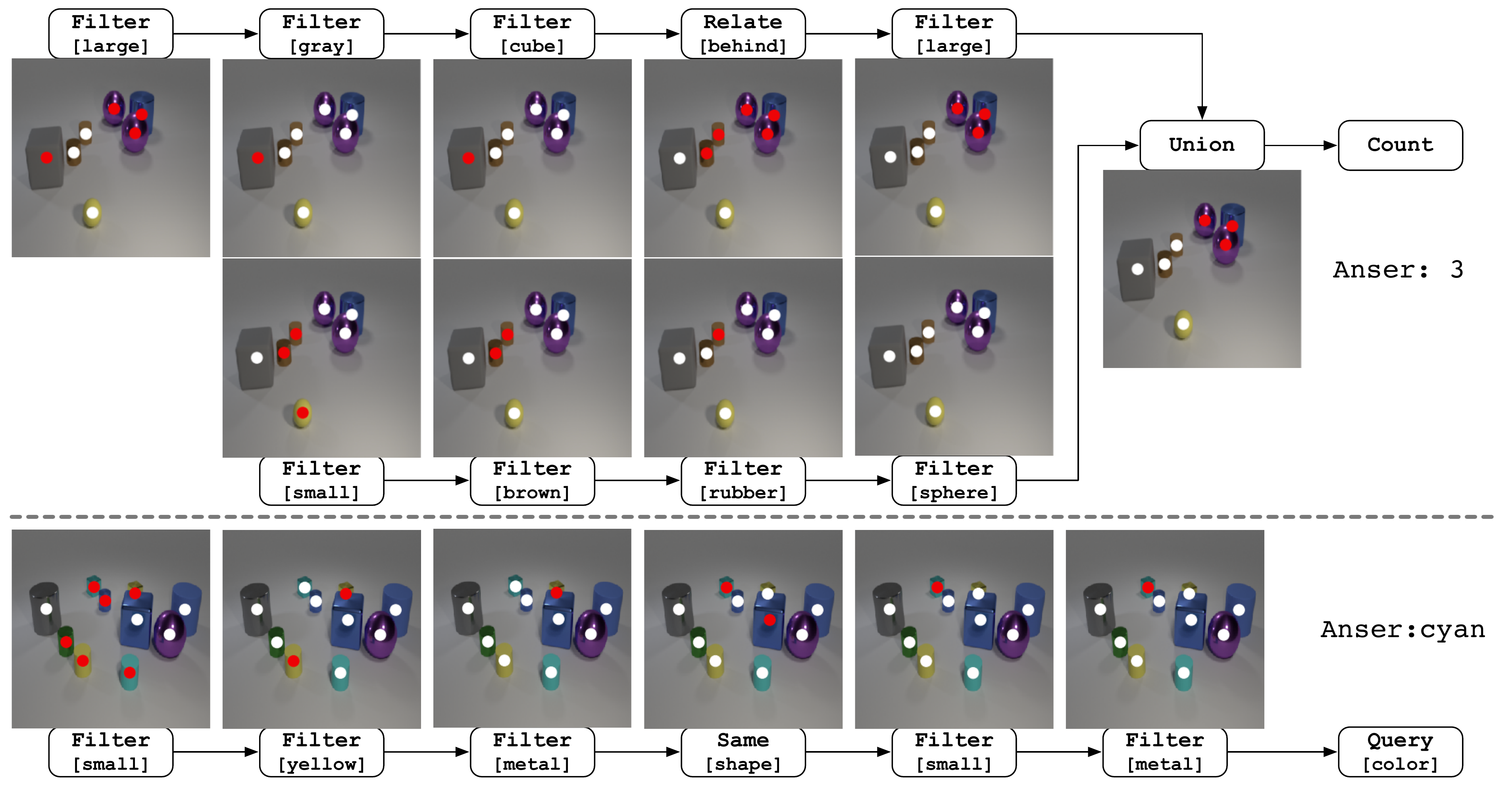}
\caption{Reasoning visualizations of two CLEVR samples. Question 1: \textit{What number of objects are either big objects that are behind the big gray block or tiny brown rubber balls?} Question 2: \textit{The other small shiny thing that is the same shape as the tiny yellow shiny object is what color?} We plot a dot for each object and darker (red) dots indicate higher attention weights.}
\label{fig:vis}
\vspace{-0.3cm}
\end{figure*}

\textbf{A4:}
As our XNMs are attention-based, the reasoning process is totally transparent and we can easily show intermediate results.
Figure~\ref{fig:vis} displays two examples of CLEVR.
We can see all reasoning steps are clear and intuitive.

\subsection{CLEVR-CoGenT}

\begin{table}[h]
    \caption{Comparisons between NMNs on CLEVR-CoGenT. Top section: results of the test set; Bottom section: results of the validation set. Using the ground-truth scene graphs, our XNMs generalize very well and do not suffer from shortcuts at all.}
    \centering
    \small
    \begin{tabular}{|c|c|c|c|}
    \hline
    Method                          &    Program   &    Condition A     &   Condition B \\
    \hline
    PG+EE~\cite{johnson2017inferring} &     supervised    &     96.6    &   73.7    \\
    TbD-net~\cite{mascharka2018transparency} &     supervised    &   \textbf{98.8}    &    \textbf{75.4} \\
    XNM-Det  &     supervised    &    98.1   &   72.6 \\
    \hline
    \hline
    NS-VQA~\cite{yi2018nsvqa}       &     supervised    &    99.8    &    63.9 \\
    XNM-Det  &     supervised    &    98.2   &   72.1 \\
    XNM-Det  &     GT    &    98.3   &   72.2 \\
    XNM-GT  &     supervised    &    99.9   &   99.9  \\
    XNM-GT  &     GT    &    \textbf{100}   &   \textbf{100}  \\
    \hline
    \end{tabular}
    \label{tab:cogent-results}
\vspace{-0.3cm}
\end{table}

\textbf{Settings.}
The CLEVR-CoGenT dataset is a benchmark to study the ability of models to recognize novel combinations of attributes at test-time, which is derived from CLEVR but has two different conditions:
in Condition A all cubes are colored one of gray, blue, brown, or yellow, and all cylinders are one of red, green, purple, or cyan; in Condition B the color palettes are swapped.
The model is trained using the training set of Condition A, and then is tested using Condition B to check whether it can generalize well to the novel attribute combinations.
We train our model on the training set of Condition A, and report the accuracy of both conditions.

\textbf{Goals.}
\textbf{Q1:}
Can our model perform well when meeting the novel attribute combinations? 
\textbf{Q2:}
If not, what actually causes the reasoning shortcut?

\textbf{Results.}
Results of CLEVR-CoGenT are displayed in Table~\ref{tab:cogent-results}.
\textbf{A1:}
When using the ground-truth scene graphs, our XNMs perform perfectly on both Condition A and Condition B.
Novel combinations of attributes in Condition B do not cause the performance drop at all.
However, when using the detected scene graphs, where node embeddings are RoI features that fuse all attribute values, our generalization results on Condition B drops to 72.1\%, suffering from the dataset shortcut just like other existing models~\cite{johnson2017inferring,mascharka2018transparency}.

\begin{figure}[H]
\includegraphics[width=\linewidth]{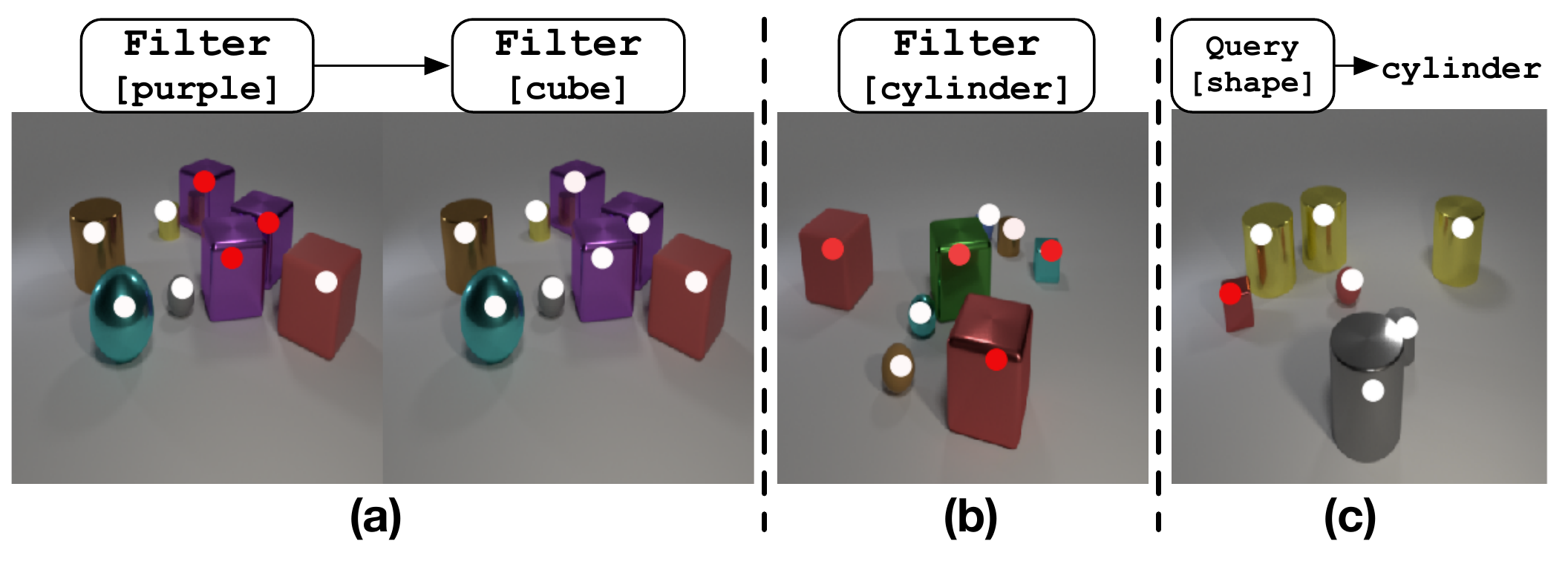}
\caption{Failure cases of our Det setting on Condition B of CoGenT.}
\label{fig:fail}
\vspace{-0.2cm}
\end{figure}

\textbf{A2:}
Figure~\ref{fig:fail} shows some typical failure cases of our Det setting on Condition B.
In case (a), our model cannot recognize purple cubes as ``cube'' because all cubes are colored one of gray, blue, brown, or yellow in the training data.
Similarly, in case (b) and (c), whether an object is recognized as a ``cube'' or a ``cylinder'' by our model is actually determined by its color.
However, in our GT setting, which is given the ground-truth visual labels, we can achieve a perfect performance.
This gap reveals that the challenge of CLEVR-CoGenT mostly comes from the vision bias, rather than the reasoning shortcut.

\subsection{VQAv2.0}
\textbf{Settings.}
VQAv2.0~\cite{goyal2017vqa2} is a real-world visual Q\&A dataset which does not have annotations about scene graphs and module programs.
We used the grounded visual features of \cite{anderson2018bottom} as node features, and concatenated node embeddings as edge features.
We set $K=1$ and fused the question embedding with our output feature for answer prediction.
Following \cite{anderson2018bottom}, we used softmax over objects for node attention computation.

\textbf{Goals.}
We used VQAv2.0 to demonstrate the generality and robustness of our model in the practical case.

\textbf{Results.}
We list the results in Table~\ref{tab:vqa-results}.
We follow StackNMN~\cite{hu2018explainable} to build the module program in a stacked soft manner, but our model can achieve better performance as our reasoning over scene graphs is more powerful than their pixel-level operations.

Recall that~\cite{johnson2017inferring,mascharka2018transparency} are not applicable in open-vocabulary input, and~\cite{yi2018nsvqa} relies on the fixed label representation, so it is hard to apply them on practical datasets.
In contrast, our XNMs are flexible enough for different cases.

\begin{table}[h]
    \caption{Single-model results on VQAv2.0 validation set and test set. $\dagger$: values reported in the original papers.}
    \centering
    \footnotesize
    \begin{tabular}{|c|c|c|c|}
    \hline
    Method                          &    expert layout   &   validation(\%)   &   test(\%)  \\
    \hline
    Up-Down~\cite{anderson2018bottom}   &   no   &   63.2$^\dagger$   &  66.3 \\
    N2NMN~\cite{hu2017learning}    &    yes      &     -    &  63.3$^\dagger$   \\
    StackNMN~\cite{hu2018explainable}  &    no   &     -    &  64.1$^\dagger$  \\
    XNMs     &     no     &  \textbf{64.7}   &   \textbf{67.5}    \\
    \hline
    \end{tabular}
    \label{tab:vqa-results}
\vspace{-0.4cm}
\end{table}

\section{Conclusions}
In this paper, we proposed X neural modules (XNMs) that allows visual reasoning over scene graphs, represented by different detection qualities.
Using the ground-truth scene graphs and programs on CLEVR, we can achieve 100\% accuracy with only 0.22M parameters. Compared to existing neural module networks, XNMs disentangle the ``high-level'' reasoning from the ``low-level'' visual perception, and allow us to pay more attention to teaching A.I. how to ``think'', regardless of what they ``look''. We believe that this is an inspiring direction towards explainable machine reasoning. Besides, our experimental results suggest that visual reasoning benefits a lot from high-quality scene graphs, revealing the practical significance of the scene graph research.

\textbf{Acknowledgments.}
The work is supported by NSFC key projects (U1736204, 61661146007, 61533018), Ministry of Education and China Mobile Research Fund (No. 20181770250), THUNUS NExT Co-Lab, and Alibaba-NTU JRI.

{\small
\bibliographystyle{ieee}
\bibliography{egbib}
}

\newpage
\appendix
\section*{Appendix}
\section{Implementation Details}
In both of the GT and Det experiments on the CLEVR dataset, we set the dimension of the label embedding (\ie, $d$) and the dimension of $\mathbf{h}$ in all output modules to $128$. The classifier consists of a simple multi-layer perceptron that maps the $128$-dimentional features to the number of possible answers (\ie, $28$), and a softmax layer.
We used Adam~\cite{kingma2014adam} optimizer with an initial learning rate $0.001$ to train our module parameters. We trained for $5$ epochs for the GT setting and $10$ epochs for the Det setting, and we reduced the learning rate to $0.0001$ after the first epoch. Each epoch takes about one hour with an Nvidia 1080Ti graphic card.

The mapping matrices of the \texttt{Describe} module are implemented differently between the GT and Det setting.
In the GT setting, as each object has four attribute values corresponding to four attribute categories (\ie, color, shape, size, material), and our node embedding is the concatenation of attribute label embeddings, the dimension of node embeddings is $4d$, where $d$ is the dimension of label embeddings.
We fix the order of label embedding concatenation as [color, shape, size, material], so we can extract node $i$'s color feature by $[\mathbf{I}_d; \mathbf{0}_d; \mathbf{0}_d; \mathbf{0}_d] \cdot \mathbf{v}_i$, where $\mathbf{I}_d, \mathbf{0}_d$ are $d \times d$ identity matrix and zero matrix respectively.
So in the GT setting, our four mapping matrixes are defines as:
\begin{equation}
\begin{aligned}
    \mathbf{M}_1 = [\mathbf{I}_d; \mathbf{0}_d; \mathbf{0}_d; \mathbf{0}_d],\\
    \mathbf{M}_2 = [\mathbf{0}_d; \mathbf{I}_d; \mathbf{0}_d; \mathbf{0}_d],\\
    \mathbf{M}_3 = [\mathbf{0}_d; \mathbf{0}_d; \mathbf{I}_d; \mathbf{0}_d],\\
    \mathbf{M}_4 = [\mathbf{0}_d; \mathbf{0}_d; \mathbf{0}_d; \mathbf{I}_d].\\
\end{aligned}
\end{equation}
In the Det setting, we regard $\mathbf{M}_k \in \mathbb{R}^{d \times d}$ as parameters and learn them automatically, which leads to an increase of the number of parameters (Det has 0.55M parameters while Gt only has 0.22M).

As for the attention functions in the CLEVR GT experiments, besides the label-space softmax which is mentioned in Eq.~\ref{eq:softmax-att}, we have also tried the sigmoid activation. Specifically, we fused multiple label vectors into one vector via a fully connected layer, and then applied the sigmoid function like Eq.~\ref{eq:sigmoid-att} to separately compute attention weights of each node and edge. Using this attention strategy, we can still obtain 100\% accuracy in the GT setting, demonstrating that our model is robust and flexible.

In the VQAv2.0 dataset, we selected the most frequent 3000 answers from the training set, and predicted the target answer from these candidates.
We used an LSTM as the question encoder and fused the $1024$-dimensional question embedding with the output feature from our module network for the answer classification.
For the training, we used Adam optimizer with an initial learning rate $0.0008$ and we set batch size as $256$.
The learning rate was decayed by half every $50000$ training steps and the training lasted for $100$ epochs.

\section{Failure Cases of the CLEVR Det Setting}
\begin{figure*}[ht]
    \centering
    \includegraphics[width=\linewidth]{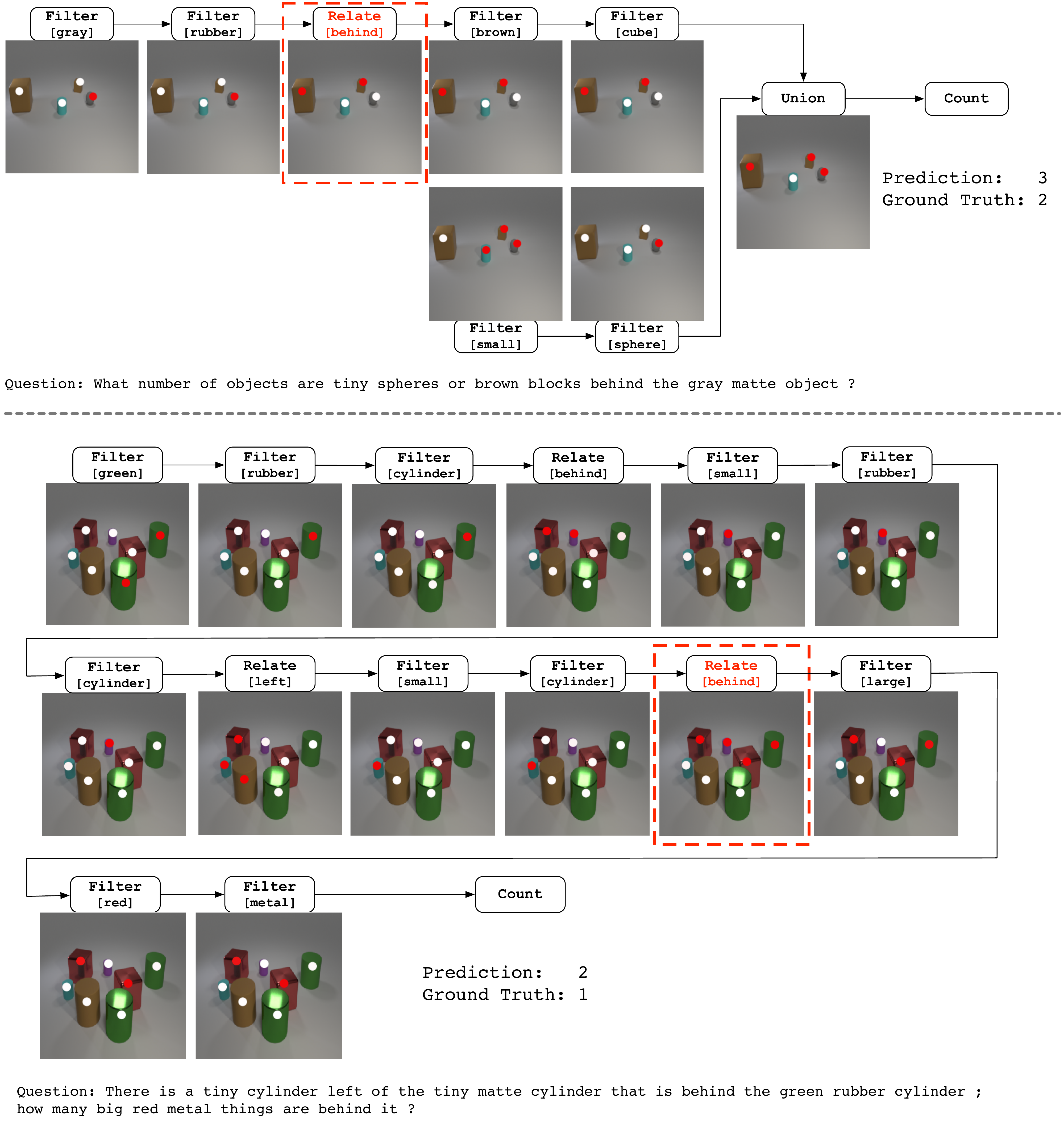}
    \caption{Failure cases caused by the inaccurate coordinate detection. Top case: the large brown cube is not behind the gray rubber object. Bottom case: a large red cube is wrongly recognized to be behind the tiny cylinder.}
    \label{fig:case1}
\end{figure*}

\begin{figure*}[ht]
    \centering
    \includegraphics[width=\linewidth]{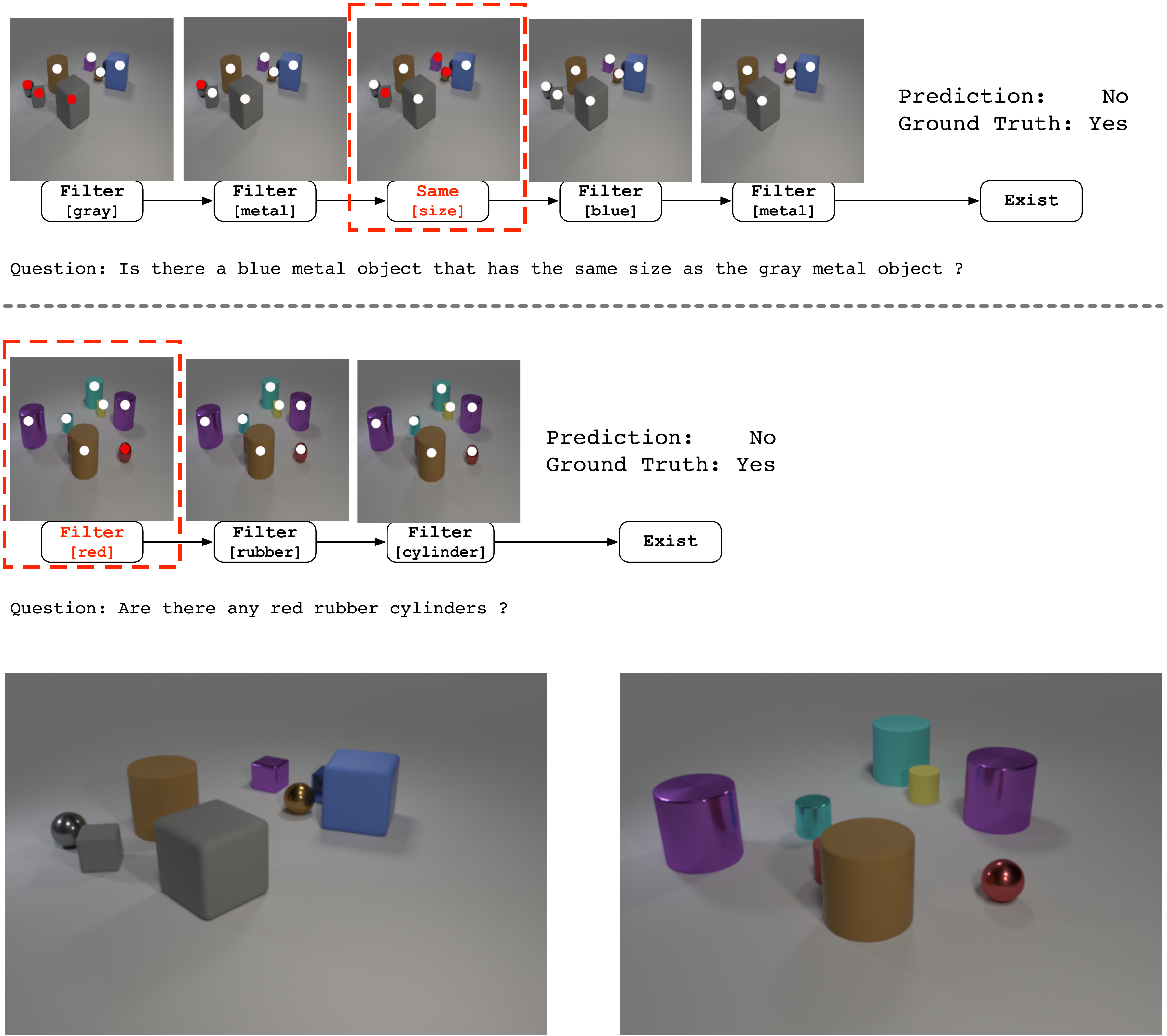}
    \caption{Failure cases caused by occluded objects. We show the high-resolution images here, and we can see that 1) in the left image, there is a small blue cube behind the large blue cube occluded; 2) in the right image, there is a red cylinder behind the large brown cylinder occluded. These occluded objects do not have corresponding dots in the reasoning results, leading to a wrong prediction ``No'' while the actual answer is ``Yes''.}
    \label{fig:case2}
\end{figure*}

\begin{figure*}[ht]
    \centering
    \includegraphics[width=\linewidth]{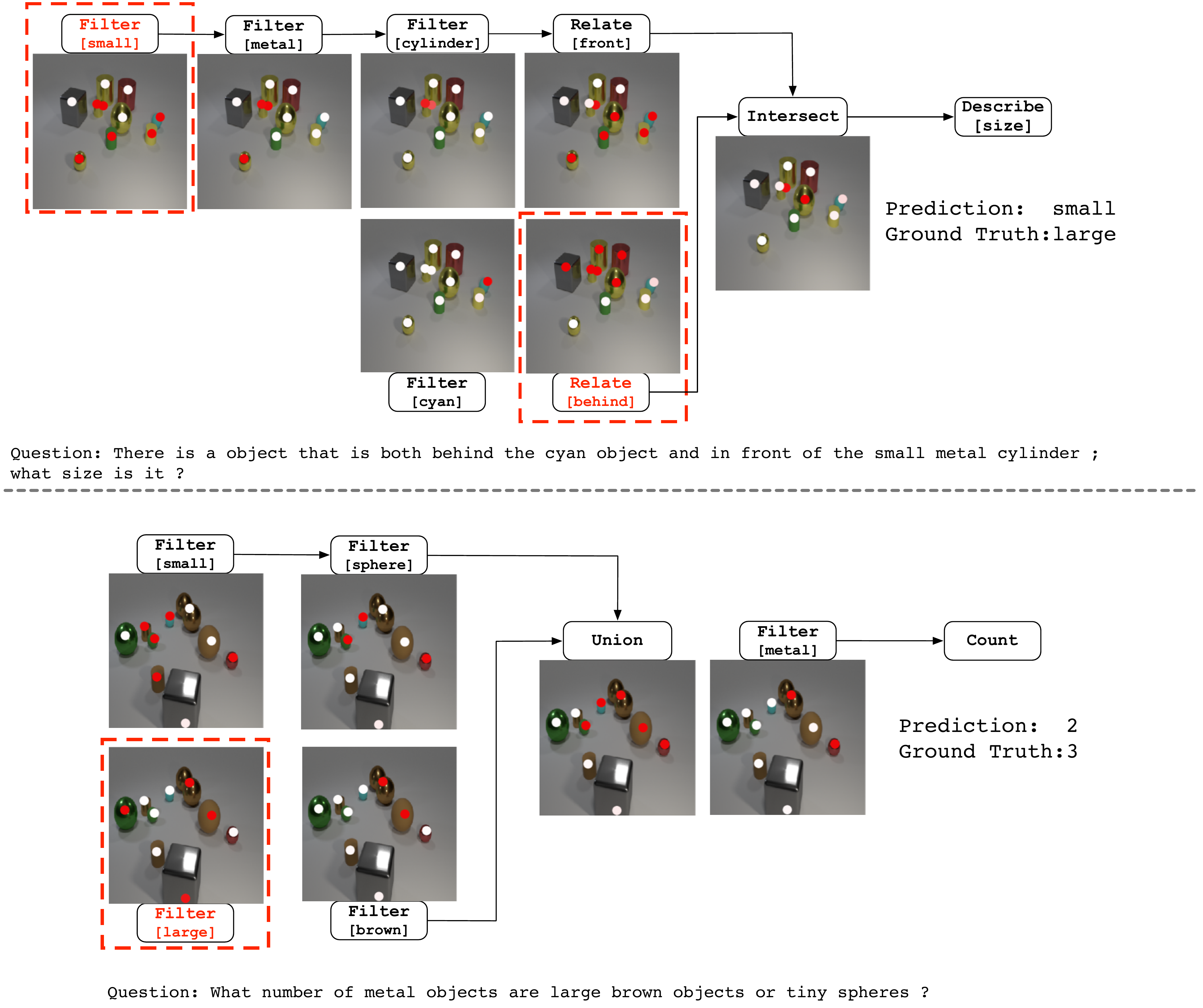}
    \caption{Failure cases caused by the inaccurate object division. Top case: two dots are assigned to the same object. Bottom case: two adjacent objects with the same attribute values (\ie, large, brown, sphere, metal) are recognized as one object, which makes the predicted number less than the ground truth answer.}
    \label{fig:case3}
\end{figure*}

We classify the failure cases in the CLEVR Det setting into three categories:
\begin{enumerate}
    \item The coordinate detection is inaccurate (Figure~\ref{fig:case1}).
    \item Some objects are occluded (Figure~\ref{fig:case2}).
    \item The mask-based object division is inaccurate (Figure~\ref{fig:case3}). Specifically, when we propose objects based on the generated mask from the ``Attend'' modules of \cite{mascharka2018transparency}, we may wrongly propose more or less objects due to the blurring boundaries.
\end{enumerate}
In all of Figure~\ref{fig:case1}, \ref{fig:case2}, and \ref{fig:case3}, \textbf{we mark the reasoning steps that cause mistakes in red box}.
We can see that using our X reasoning over scene graphs, we can easily track the reasoning process and diagnose where and why the model makes a mistake, which is an inspiring step towards the explainable AI.

\section{Case Study on the VQAv2.0}
\begin{figure*}[ht]
    \centering
    \includegraphics[width=\linewidth]{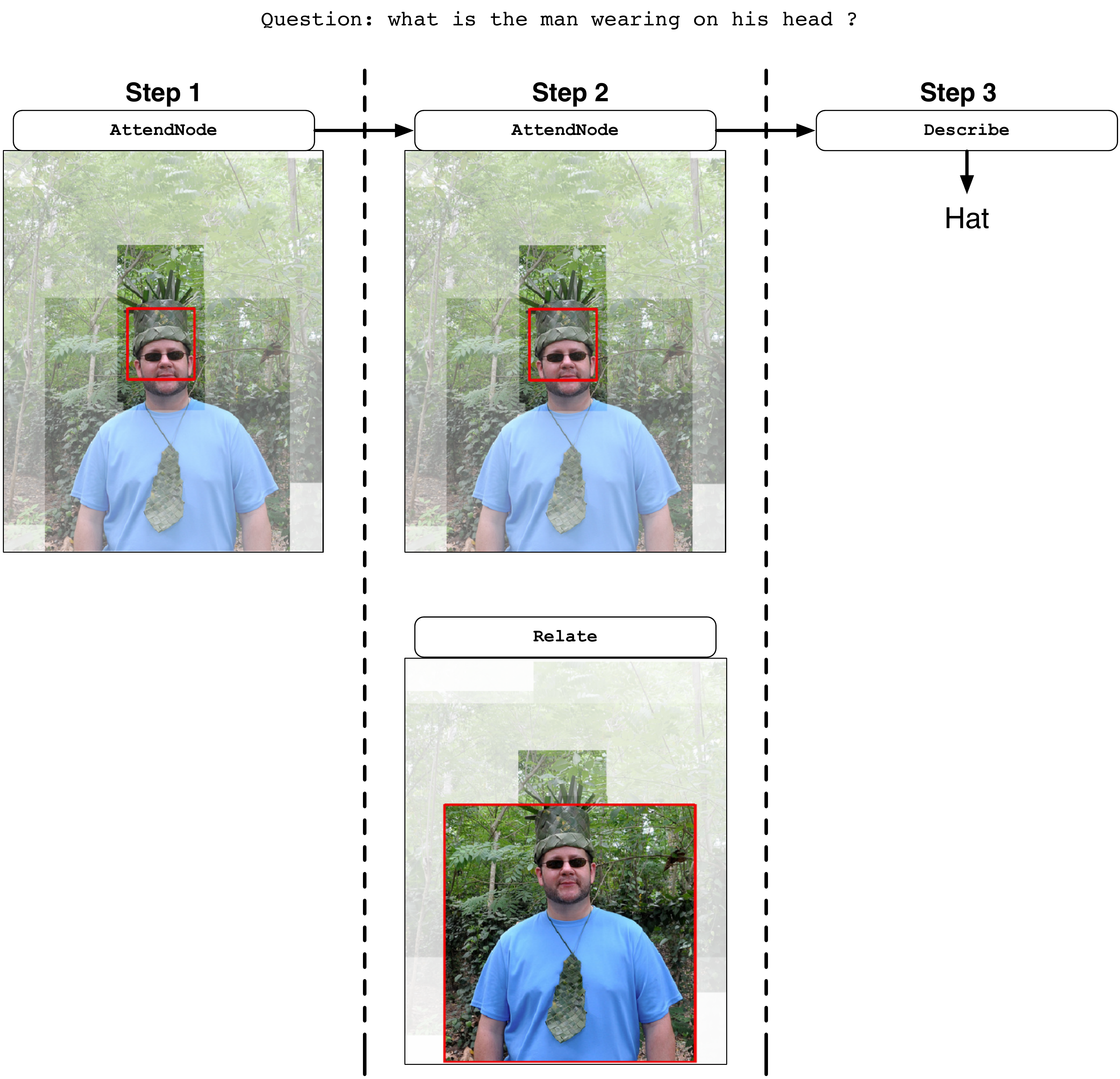}
    \caption{A typical case of the VQAv2.0 dataset. The top row lists the modules with the maximum probability at each step. We can see that even the question ``what is the man wearing on his head'' explicitly requires the relationship understanding, the module \texttt{Relate} is still not necessary as the target region can be directly found. We argue the simplicity of question annotations is a major shortcoming of the VQAv2.0 dataset.}
    \label{fig:case_vqa}
\end{figure*}

In the VQAv2.0 experiments, we predict a probability distribution over our modules at each step, and then feed the soft fusion of their outputs into next step.
We set the reasoning length to $3$ and force the last module to be \texttt{Describe}.
We show a typical sample of the VQAv2.0 dataset in Figure~\ref{fig:case_vqa}.
The top row is the modules with the most probability at each step, while the bottom row shows the results of \texttt{Relate} at Step 2.
We can see that even though the question ``what is the man wearing on his head'' explicitly requires the relationship reasoning (\ie, find the ``man'' first, and then move the focus to his head), our model scarcely relies on the results of the \texttt{Relate} module.
Instead, it can directly focus on the ``head'' and give the correct prediction.
We think this is due to the simplicity of questions, which is a shortcoming of the VQAv2.0 dataset.

\end{document}